\documentclass[runningheads]{llncs}

\usepackage{eccv}

\usepackage{eccvabbrv}

\usepackage{graphicx}
\usepackage{booktabs}
\usepackage{todonotes}
\usepackage{graphicx}
\usepackage{multirow}
\usepackage{longtable}

\usepackage[accsupp]{axessibility}  

\usepackage{hyperref}

\usepackage{orcidlink}

\usepackage{comment}
\usepackage{float}

\usepackage{pifont}
\newcommand{\cmark}{\ding{51}}
\newcommand{\xmark}{\ding{55}}

\begin{document}

\title{BenthicFlow: Generating Extensible Underwater Environments via Flow Matching} 

\titlerunning{BenthicFlow: Extensible Underwater Environments}

\author{Joaquín Figueira\textsuperscript{*} \and 
Camile Lendering\textsuperscript{*} \and
Manfred Gonzalez-Hernandez \and
Giacomo D'Amicantonio \and
Erkut Akdag \and
Egor Bondarev}

\authorrunning{J.~Figueira, C.~Lendering et al.}

\institute{
AIMS Group, Department of Electrical Engineering, Eindhoven University of Technology,
Eindhoven, The Netherlands\\
\email{\{j.figueira,c.r.lendering\}@tue.nl}\\[2pt]
\textsuperscript{*}Equal contribution.
}

\maketitle

\begin{abstract}
  Computer vision applications for 3D scene understanding in underwater environments remain challenging due to the lack of high-quality 3D data and the inability of surface-trained models to generalize to underwater scenes. To address this challenge, an emerging trend is to employ generative models to close the data domain gap. However, existing methods assemble large scenes by stitching independently generated tiles post hoc with separately trained models, while demonstrating heterogeneous landscapes only within individual survey sites. We introduce BenthicFlow, a unified framework based on a single conditional flow-matching model that jointly generates aligned textures and depth maps. A MultiDiffusion-inspired sampling procedure reconciles overlapping windows throughout the generative trajectory, enabling spatially extensible RGBD mosaics without a separate stitching model. The generated mosaics are subsequently lifted into explicit 3D benthic environments using surface-aligned Gaussian surfels. Experiments across geographically distinct survey sites demonstrate that BenthicFlow preserves site-specific appearance while generating coherent, large-scale 3D scenes that closely match the target distributions. Code and trained models are available at \url{https://github.com/jacomof/BenthicFlow}.

\end{abstract}

\vspace{-15pt}

\section{Introduction}
\label{sec:intro}

The ocean floor ranks among the least explored surfaces on Earth~\cite{Grosvik_status_2023}, yet autonomous underwater vehicles (AUVs) have quietly accumulated vast image collections over the past two decades~\cite{Zurowietz_MAIA_A_2018}. A single modern deployment routinely returns tens of thousands of nadir RGB frames, covering square-kilometer tracts of reef, sand, and rubble in one mission. Consequently, data acquisition is no longer the primary bottleneck; scene interpretation is. However, these images are collected without captions, reliable camera poses, and dense 3D supervision. In addition, low-texture seafloor and drifting particulate sediment routinely break structure-from-motion (SfM) and simultaneous localization \& mapping (SLAM), while active depth \& light-detection-and-ranging (LiDAR) sensor data severely degrade in the water column~\cite{underwater_lidar}. In summary, raw data is abundant; usable 3D structured data is scarce.

A generative model of the seafloor would narrow this gap for several communities.
In marine robotics applications~\cite{Choi_PhysicsRobotics_2021,Mai_SemanticSegRobotics_2025, Song_OceanSim_2025}, large-scale synthetic terrain could populate the simulators
that train and stress-test perception systems ahead of a mission, precisely where
target-site data cannot be gathered in advance. In marine ecology scenarios~\cite{Grosvik_status_2023,Zurowietz_MAIA_A_2018}, a model of the
distribution of healthy seafloor could act as a normality prior, against which
anomalies such as bleaching, invasive species, or debris are flagged as
outliers, without any curated set of negative examples. Underwater scenes also
remain underrepresented in foundation-model training data~\cite{Zheng_MarineInst_2024}, which is
dominated by terrestrial, human-centric imagery. Prompted for the seafloor,
general-purpose text-to-image models default to a stylized ocean of saturated
coral and clear blue water rather than the flat, turbid, nadir-view frames that
survey platforms record. The mismatch stems from the training distribution rather than the model's aesthetic quality, and
it is precisely what a normality prior must capture, since a sample that is
plausibly marine yet distributionally wrong provides no reliable reference. A
model trained directly on survey imagery is therefore required.

The central distinction of our work relative to similar underwater generative
pipelines lies in how large scenes are assembled. The closest method, the
extensible-terrain generator of~\cite{zhang_infinite_2025}, conditions a denoising
diffusion model on foundation-model embeddings and assembles large maps from
independently sampled tiles whose coherence is established only after sampling.
That strategy carries three key limitations. First, it requires training a separate unconditional
in-painting model to blend neighboring tiles, as
conditional in-painting is reported to introduce boundary artifacts. Second, because seam regions are synthesized independently of the conditioning process, they are generated without access to the latent representations that guide the appearance of the remaining scene. Third, consistency is enforced only between neighboring tile pairs rather than jointly across the entire scene, limiting global coherence. Moreover, every appearance
transition demonstrated in that work remains within a single survey site, whereas generation across geographically distinct sites is not shown.

To address these limitations, we propose a generative pipeline termed \textit{BenthicFlow}. The pipeline jointly models appearance and geometry by combining continuous-time flow matching and monocular depth estimation, leveraging 
marine imagery from the Squidle+ collaborative data
framework~\cite{friedman2025squidle}. A single conditional flow-matching model is trained to synthesize aligned RGB-depth (RGBD) pairs. These pairs are extended into
RGBD mosaics of unbounded spatial extent using a windowed sampling strategy, inspired
by MultiDiffusion~\cite{bar2023multidiffusion}, where generation is initialized from reference seeds distributed across a plane. The resulting mosaics are projected into point clouds and
reconstructed into a spatially continuous 3D environment via Gaussian splatting,
yielding realistic benthic scenes. The proposed architecture addresses the limitations of existing underwater generative pipelines through its unified formulation. Because a single flow-matching model treats generation and in-painting
as one operation, overlapping windows are reconciled by averaging their predicted
velocities during sampling rather than being patched together afterward. No
secondary stitching network is required, and all regions of the canvas, including seam bands,
are synthesized under the same latent conditioning. Furthermore, coherence is imposed
jointly across all overlapping windows at every integration step, enabling scalable scene generation within a single shared sampling trajectory. Finally, a single conditional velocity field models multiple geographically distinct survey sites and interpolates between them, a capability that the prior underwater generators have not demonstrated. To summarize, the main contributions of this work are as follows:
\begin{itemize}
    \item A windowed flow-matching formulation for extensible benthic substrate generation, in which
    overlapping latent windows are reconciled throughout a shared generative
    trajectory. This enables spatially extensible RGBD mosaics using a single
    generative model, without a separately trained stitching or in-painting network.
    \item A single conditional velocity field that spans multiple geographically
    distinct benthic survey sites and interpolates smoothly between them.
    \item A lifting stage that converts RGBD benthic mosaics into spatially continuous 3D representations using surface-aligned Gaussian splats.
\end{itemize}

\section{Related Works}
\label{sec:related-work}

\subsection{Marine Computer Vision}

Underwater computer vision spans numerous applications, including marine biology~\cite{Piechaud2022mapping, zhong2023combining, piechaud2019automated}, infrastructure inspection~\cite{jiao2024vision, ma2023rov}, autonomous underwater vehicle (AUV) navigation~\cite{chang2022active, manzanilla2019autonomous}, mapping~\cite{zhong2024application, wang2023real}, and marine archaeology~\cite{van2015computer, johnson2017high}. Due to unique optical degradation, a large portion of the literature focuses on image restoration. Methods range from combining deep learning with standard optics, such as SeaThru~\cite{akkaynak_sea-thru_2019}, to end-to-end approaches such as Semi-UIR~\cite{huang_contrastive_2023}. Recent works, such as DPF-Net~\cite{mei_dpf-net_2025}, incorporate monocular depth estimation to jointly correct distortions and physical light properties. Building on these optical models, 3D reconstruction techniques have adapted Neural Radiance Fields to scattering media, as seen in SeaThru-NeRF~\cite{levy_seathru-nerf_2023}, with subsequent refinements for dynamic scenes~\cite{tang_neural_2024} and improved processing efficiency~\cite{li_watersplatting_2025, hou_real-time_2022}.

In contrast, generating synthetic underwater data typically serves downstream tasks. Many models focus on surface-to-underwater image transfer by learning optical characteristics~\cite{wu_self-supervised_2024, Ye_2022_CVPR, desai2021ruig}, while others like UW-GAN~\cite{hambarde2021} generate data to extract accurate depth estimates as a byproduct. Most closely related to our approach is \cite{zhang_infinite_2025}, which
similarly couples joint image and depth generation with differentiable
rendering, and likewise steers appearance through a spatial field of
foundation-model latents. The two methods diverge in how scenes reach
arbitrary sizes: in~\cite{zhang_infinite_2025}, the tiles are sampled independently and reconciled after
sampling by a separately trained inpainting model
(RePaint~\cite{lugmayr2022repaint}), whereas our sampler resolves overlapping
windows jointly within the generative trajectory itself. Traditionally, generating expansive environments relies on procedural modeling, utilizing layered noise functions and physical simulations~\cite{galin2019review}. Platforms like HoloOcean~\cite{potokar2022holoocean} deploy these deterministic assets for autonomous vehicle benchmarking. However, procedural synthesis struggles to replicate the organic variations and complex light-scattering of real marine biomes~\cite{zhang_infinite_2025}. Consequently, our framework diverts from hard-coded scripts, replacing them with a fully learned approach to capture authentic topologies and textures.

\subsection{Continuous Generative Modeling and Latent Representations}
Generative modeling has increasingly adopted continuous-time Flow Matching (FM)~\cite{lipman2023flow, liu2022flow, albergo2023building}, whose tractable conditional formulation (CFM) regresses a velocity field onto per-sample conditional paths, admitting near-straight flow-matching trajectories and competitive sample quality at fewer function evaluations than denoising diffusion models. Scaling has established FM as the state of the art for image synthesis, e.g., Qwen-Image~\cite{wu2025qwenimagetechnicalreport}, FLUX.2~\cite{bflFLUX2Frontier}, and rectified flow transformers~\cite{esser2024scaling}.

These models typically operate in compressed latent spaces to reduce computational costs. Traditional frameworks rely on VQ-GANs~\cite{esser2021taming} or Variational Autoencoders (VAEs)~\cite{rombach2022high}, but their strict information bottlenecks often produce highly non-linear distributions that complicate velocity estimation. Representation Autoencoders (RAEs)~\cite{zheng2026diffusion} address this by mapping data into a high-dimensional semantic space derived from frozen Vision Foundation Models, easing the learning dynamics for downstream Transformers. We build on RAEs, extending them to joint multi-modal (RGBD) tokens as the input space for our conditional flow-matching model.

 Finally, to generate expansive structures beyond standard training resolutions, our approach leverages spatial blending. This paradigm was introduced by MultiDiffusion~\cite{bar2023multidiffusion}, a training-free framework that unifies localized generation paths by averaging overlapping crops during sampling. Concurrent work~\cite{jimenez2023mixturediffusersscenecomposition} refined this using a Gaussian-weighted averaging strategy that prioritizes window centers. We extend these fusion mechanisms into continuous flow trajectories, combining multi-window generation directly within the velocity estimation steps.

\section{Dataset}
\label{sec:dataset}

The dataset in this work comprises over 600k benthic nadir frames from the Squidle+ framework~\cite{friedman2025squidle}, collected by an AUV across two Australian reefs and the northwestern coast of Hawaii. Capturing diverse marine fauna, flora, substrates, and anthropogenic debris, all frames are color-normalized and standardized to $518\times518$ pixels. The data is organized hierarchically into campaigns (encompassing a specific location) and individual deployments (sequential images from a continuous time frame). To prevent temporal data leakage, we partition the dataset strictly at the deployment level. For each campaign, one deployment is reserved for validation ($\approx$90k images), one for testing ($\approx$90k images), and the remainder for training ($\approx$500k images). In the supplementary material we detail the specific campaigns and deployments selected for each data split.

Since standard underwater surveys lack dense geometry, a per-pixel depth map is estimated and retained as a fourth channel. Furthermore, to mitigate optical degradations like barrel distortion and non-uniform illumination, we evaluate two distinct data preprocessing pipelines. The first applies Contrast Limited Adaptive Histogram Equalization (CLAHE) for illumination normalization coupled with relative depth estimated by Depth Anything V2~\cite{yang2024depth} (DAv2), forming our \textit{BenthicFlow} configuration. The second model, \textit{BenthicFlow-DPF}, uses the same source data and architecture, but leverages DPF-Net~\cite{mei_dpf-net_2025} to provide both image enhancement pre-processing and produce estimated depth.

\section{Methodology}
\label{sec:methodology}

\begin{figure}
    \centering
    \includegraphics[width=1.0\linewidth]{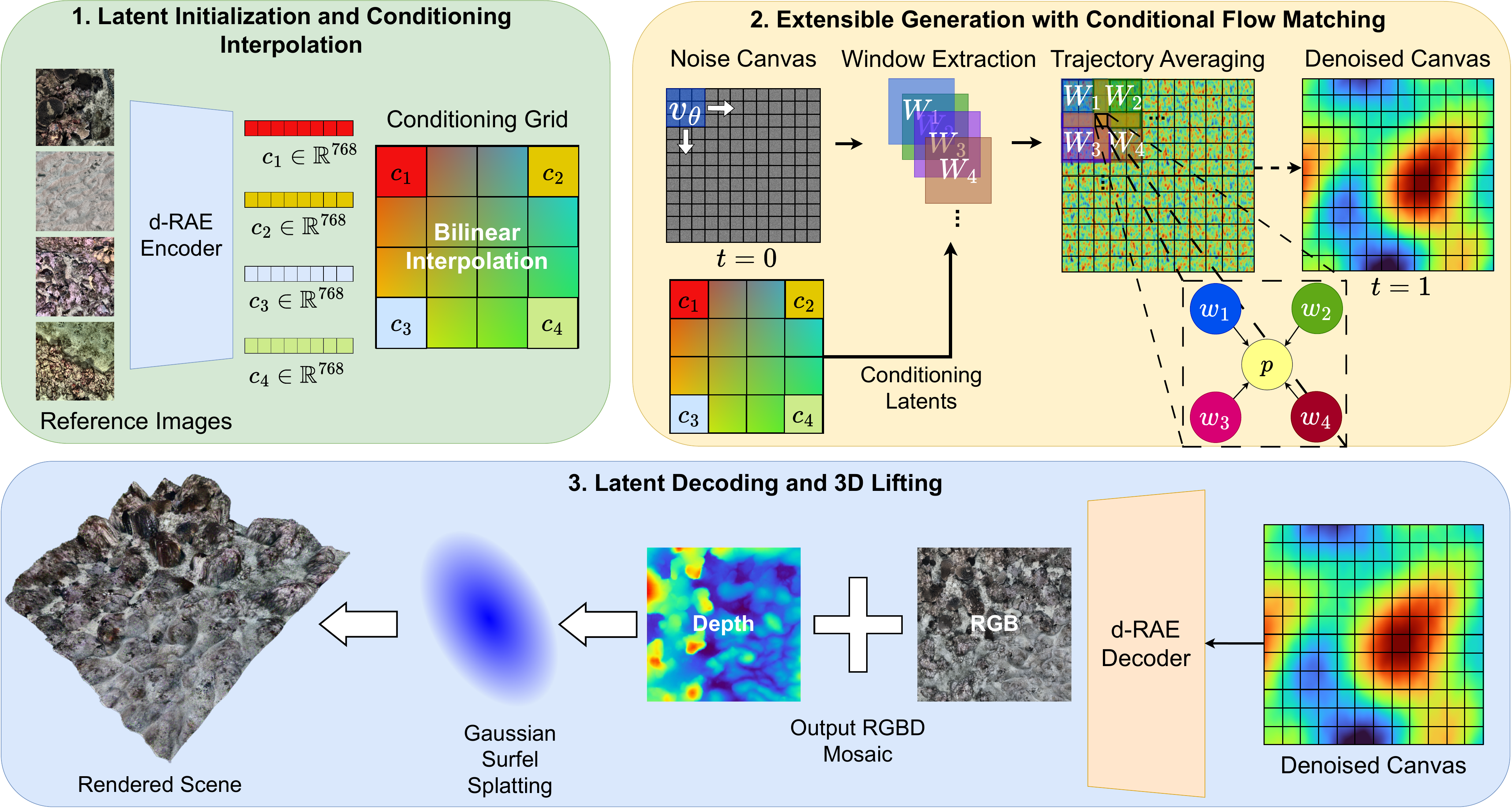}
    \caption{Overview of BenthicFlow's generative pipeline. First, reference images are encoded and interpolated to form a large conditioning embedding grid. A large Gaussian noise canvas is then sampled and convolved in overlapping windows by a CFM model ($v_{\theta}$) trained to denoise small latent windows ($W_i$). This process is repeated over multiple integration steps, and, on each step, the flow trajectories of overlapping tokens are averaged. Finally, the complete denoised canvas is decoded in a single forward pass into a large RGBD mosaic, which is splatted to render the final scene.}
    \label{fig:extensible-generation-diagram}
    \vspace{-10pt}
\end{figure}

In this section, we introduce our BenthicFlow generative pipeline, which incorporates elements from conditional flow matching, multi-diffusion sampling, and gaussian splatting into a flexible underwater terrain generator depicted in \cref{fig:extensible-generation-diagram}. To maintain computational feasibility,  BenthicFlow operates within a compressed latent space managed by a modified formulation of the  Representation Autoencoder (RAE)~\cite{zheng2026diffusion}, which jointly encodes images and depth maps. BenthicFlow's autoencoder, named d-RAE, consists of a frozen DINOv2 backbone used to process RGB images, coupled with a separate ViT trained from scratch to produce depth map latent representations. Formally, given an image $I \in \mathbb{R}^{3 \times H \times W}$ and its corresponding depth map $D \in \mathbb{R}^{1 \times H \times W}$, d-RAE encodes them into a fused latent representation:
\begin{equation}
\label{eq:encoding-equation}
    z = \mathcal{N}_{\text{tok}}\big(\big[E_{\text{rgb}}(I) \,;\, E_{\text{d}}(D)\big]\big) \in \mathbb{R}^{h \times w \times 1024},
\end{equation}
where $(h,w) = (H/14, W/14)$. d-RAE's RGB encoder  ($E_{\text{rgb}}$) encodes the images into patch tokens $E_{\text{rgb}}(I) \in \mathbb{R}^{h \times w \times 768}$, while the ViT depth encoder ($E_{\text{d}}$) produces $E_{\text{d}}(D) \in \mathbb{R}^{h \times w \times 256}$. During training, the RGB and depth representations are aligned naturally through a reconstruction training protocol, up to a pretraining bias in the affine statistics of the DINOv2 encoder. To handle this final inconsistency, we concatenate and normalize the two sets of patch tokens by layer normalization (denoted $\mathcal{N}_{\text{tok}}$ in \cref{eq:encoding-equation}). Finally, at the decoding stage, the d-RAE model projects the latent to a width of $512$, applies a bottleneck of two residual blocks around a spatial self-attention layer, and upsamples the features to the output resolution, producing a final RGBD representation.




With this latent representation established, the overall extensible terrain generation pipeline of BenthicFlow consists of the following three stages:

\begin{enumerate}
    \item \textbf{Latent Initialization and Conditioning Interpolation:} The d-RAE's RGB encoder first compresses the input image seeds into the fused latents described above. These seed latents are subsequently mean-pooled and interpolated to form a continuous conditioning grid of a specific spatial extent (\cref{sec:latent-interpolation}).
    \item \textbf{Extensible Generation with Conditional Flow Matching:} A conditional flow matching model is convolved simultaneously across a latent canvas and the interpolated guiding grid. Spatial consistency across the canvas is maintained through flow-matching trajectory averaging during sampling. This process is repeated iteratively until the output latent has been fully denoised (\cref{sec:cfm}).
    \item \textbf{Decoding and 3D Lifting:} Finally, the fully generated latent canvas is mapped back to the pixel level by the d-RAE's fully convolutional decoder.  To produce the final scene, the reconstructed RGBD image is lifted to a continuous 3D representation through surface-aligned Gaussian surface elements (surfels) (Sec.~\ref{sec:lifting}).
\end{enumerate}

\subsection{Latent Initialization and Conditioning Interpolation}
\label{sec:latent-interpolation}

To condition BenthicFlow on specific underwater substrates, four reference images with the desired conditioning signal are processed by the d-RAE's DINOv2 encoder, obtaining 4 patch embedding representations. For each of the reference seeds $I_i$, a global appearance descriptor $c_i=\tfrac{1}{hw}\sum_{jk}E_{\text{rgb}}(I_i)_{jk}\in\mathbb{R}^{768}$ is computed. Conditioning on this pooled vector rather than a spatial map allows a campaign's benthic character to govern the generation without fixing any particular structure, enabling the downstream appearance interpolation. Since each descriptor can only condition a small portion of the image, the 4 vectors are bilinearly interpolated to the desired output scale to form a full conditioning grid, thereby letting appearance vary smoothly across the grid while the shared latent space preserves coherence.

\subsection{Extensible Generation with Conditional Flow Matching}
\label{sec:cfm}
BenthicFlow formulates the generative process as a rectified flow matching procedure between an isotropic gaussian and the d-RAE's latent representation, controlled by global appearance vectors through classifier-free guidance~\cite{ho2022classifier}. To make large scale generation feasible, a latent canvas with the target resolution is subdivided into overlapping $16\times16$ windows, each with its corresponding interpolated guiding vector. The task of BenthicFlow's CFM is to generate a single window of the latent at a time, with inter-window consistency emerging from the averaging of overlapping flow-matching trajectories.

\paragraph{Rectified flow with guidance.}
For a clean latent $z_1$ and noise $z_0\sim\mathcal{N}(0,\mathbf{I})$, the rectified
flow defines the path $z_t=(1-t)z_0+t z_1$ with target velocity $z_1-z_0$. During training, the
network $v_\theta$ regresses this velocity under uniform time steps:
\begin{equation}
    \mathcal{L}_{\text{CFM}} =
    \mathbb{E}_{t\sim\mathcal{U}(0,1),\,z_0,\,z_1,\,c}
    \big\|v_\theta(z_t,t,c)-(z_1-z_0)\big\|_2^2 .
\end{equation}
The condition is dropped to a learned null token $c_\varnothing$ with probability
$0.1$ during training. At inference, the guided velocity
$\tilde{v}=v_\theta(z_t,t,c_\varnothing)+s\big(v_\theta(z_t,t,c)-v_\theta(z_t,t,c_\varnothing)\big)$
with scale $s{=}3$ is integrated using Euler's method for ordinary differential equations approximation. Latents are standardized
per channel by statistics estimated over the training set. Note that the model is never trained explicitly on windowed denoising of large canvases. Instead, rectified flow is applied only at a fixed low resolution defined during training. 

\paragraph{Network.}
$v_\theta$ is a fully convolutional U-Net~\cite{ronneberger2015unet} over the $16\times16\times1024$ latent window,
with a single $2\times$ downsampling stage to $8\times8$, a symmetric upsampling
stage, base width $512$, and full-resolution skip connections. Each residual
block is modulated by adaptive group normalization: a sinusoidal timestep
embedding and a linear projection of a conditioning vector $c$ are summed and employed to predict per-channel scale
and shift. The output convolution is zero-initialized so that the network initially predicts a near-zero velocity field, a standard initialization for residual and generative
networks~\cite{zhang2019fixup,peebles2023dit}. An EMA copy of trainable weights is maintained for optimization stabilization.

\paragraph{Extensible Generation.}

Generation extends from one window to the full output resolution by a MultiDiffusion-inspired sampler~\cite{bar2023multidiffusion} operating directly on the
velocity field. At each Euler step, the windows are processed concurrently by $v_\theta$ and their guided velocities are averaged per token
under a separable sine window $\omega$ that downweights window borders:
\begin{equation}
    \tilde{v}_{\text{canvas}}(p)=
    \frac{\sum_{k:\,p\in W_k}\omega_k(p)\,\tilde{v}_k(p)}{\sum_{k:\,p\in W_k}\omega_k(p)} .
\end{equation}
Neighboring regions are therefore fused during sampling, rather than stitched afterward: a single model in the pipeline estimates the denoised latents, with no secondary network and no post-hoc blending, so window boundaries receive no dedicated treatment outside the sampler.

\subsection{Latent Decoding and 3D Lifting}
\label{sec:lifting}
After denoising, the frozen d-RAE decoder produces a spatially continuous RGBD mosaic from the latent canvas tokens. The fully convolutional nature of the decoder allows it to generate mosaics of arbitrary size. This represents a single 2.5D observation, which is subsequently lifted into an explicit 3D
representation for continuous novel-view rendering. Specifically, each pixel is unprojected under a pinhole model, with relative depth $d\in[0,1]$ mapped to camera range $Z=z_{\text{near}}+d\,z_{\text{span}}$, where $z_{\text{near}}$ represents the distance between the camera and the scene, and $z_{\text{span}}$ the maximum depth of any point of the terrain. Rather than isotropic 3D Gaussians, each
point is rendered as a Gaussian surface element (surfel): a per-pixel normal is estimated from the local gradient of the unprojected points, and the primitive is oriented such that its two in-plane axes span the tangent plane while a thin axis follows the normal. To keep
slanted surfaces gap-free without bridging occlusion boundaries, the surface in-plane extents are set anisotropically from the local inter-pixel spacing, measured by a one-sided minimum so a surfel never spans a depth discontinuity. Following a 2D Gaussian Splatting
formulation~\cite{huang20242d} where appearance is
observed from a single view, color is stored per primitive and view-dependent
spherical harmonics are omitted. Subsequently, an appearance-only refinement fits the source view by adjusting DC color,
opacity, and bounded in-plane scale residual. At the same time, positions, orientations, and
normal thickness stay fixed, recovering per-view fidelity without letting the
primitives inflate into blurry volumetric blobs. Optionally, a second pass polishes DC colour under an LPIPS objective while all other parameters remain fixed.

\section{Experiments}
\label{sec:experiments}

A comprehensive suite of evaluations is performed to evaluate the quality of the generative models produced in this work. The assessment is performed along three axes: single image generation and reference adherence, large-scale extensible landscape generation, and 3D benthic environment rendering.

\subsection{Model optimization}
While the BenthicFlow CFM models are trained in a single stage for 200 epochs and a batch size of 256, the d-RAEs follows a more complex three-phase curriculum lasting a total of 16 epochs and using a batch size of 16. Phase 1 applies an $\mathcal{L}_1$
reconstruction loss on RGB and depth. From epoch $6$, phase 2 adds a Learned Perceptual Image Patch Similarity (LPIPS)~\cite{zhang2018unreasonable} term
on RGB. From epoch $8$, phase 3 adds a hinge adversarial term on RGB, with a
discriminator formed from a frozen DINO-S/8 backbone and a small trainable
convolutional head, and DiffAugment applied identically to real and reconstructed
inputs. The adversarial weight is set adaptively from the ratio of reconstruction
to adversarial gradient norms at the decoder's final layer and scaled by $0.75$. Furthermore, an EMA copy of trainable weights is kept for regularization.

All models are optimized with the AdamW optimizer ($\beta{=}(0.5,0.9)$). Training inputs are $224\times224$ crops of the input images after downsampling to a $518\times518$ resolution and applying random horizontal flips. Both CFM and d-RAE models use the same EMA decay of $0.999$. The learning rate is set to $2 \times 10^{-4}$ for the d-RAE model and $1 \times 10^{-4}$ for the CFM. Training is performed utilizing 4 NVIDIA H100 GPUs, 64 CPU cores, 
and 512GB of RAM. Downstream evaluation is conducted on a single NVIDIA A100 
GPU with 16 CPU cores and 64GB of RAM.

\subsection{Single Image Generation}

\begin{table}
\centering
\caption{Quantitative comparison of image generation quality broken down by geographic region. Campaigns from Batemans Bay (3) and Scott Reef (3) are averaged to summarize regional performance, alongside Hawaii (1) (2000 images are sampled per campaign). KID is reported as the mean $\pm$ the pooled standard deviation from subset bootstrapping. BenthicFlow-DPF refers to the model using DPF-Net~\cite{mei_dpf-net_2025} pre-processing and Cos. Sim. denotes DINOv2 cosine similarity. Both BenthicFlow variants use 50 sampling steps. Best overall \textit{Average} results are highlighted in \textbf{bold}.}
\label{tab:generation_quality_regional}
\resizebox{0.8\linewidth}{!}{
\begin{tabular}{@{}ll@{\hspace{2.5em}}c@{\hspace{2.5em}}c@{\hspace{2.5em}}c@{}}
\toprule
\textbf{Model} & \textbf{Location} & \textbf{FID $\downarrow$} & \textbf{Cos. Sim. $\uparrow$} & \textbf{KID $\downarrow$} \\
\midrule
FLUX.2-dev & Batemans    & 110.3 & 0.573 & $0.0966 \pm 0.0018$ \\
        & Hawaii      & 137.1 & 0.593 & $0.1247 \pm 0.0024$ \\
                   & Scott Reef  & 76.7  & 0.745 & $0.0601 \pm 0.0014$ \\
\cmidrule(l){2-5}
                   & \textit{Average} & 99.7 & 0.649 & $0.0850 \pm 0.0017$ \\
\midrule
BenthicFlow        & Batemans    & 20.0  & 0.805 & $0.0140 \pm 0.0012$ \\
                   & Hawaii      & 13.9  & 0.833 & $0.0070 \pm 0.0005$ \\
                   & Scott Reef  & 13.2  & 0.882 & $0.0064 \pm 0.0005$ \\
\cmidrule(l){2-5}
                   & \textit{Average} & 16.2  & \textbf{0.842} & $0.0097 \pm 0.0008$ \\
\midrule
BenthicFlow-DPF    & Batemans    & 19.6  & 0.795 & $0.0119 \pm 0.0008$ \\
                   & Hawaii      & 15.5  & 0.804 & $0.0079 \pm 0.0010$ \\
                   & Scott Reef  & 11.9  & 0.894 & $0.0068 \pm 0.0005$ \\
\cmidrule(l){2-5}
                   & \textit{Average} & \textbf{15.7}  & 0.839 & \textbf{0.0091} $\pm$ \textbf{0.0007} \\
\bottomrule
\end{tabular}}
\vspace{-10pt}
\end{table}

The BenthicFlow models are evaluated against a state-of-the-art generative model that was selected due to its ability to adhere closely to reference images: 4-bit quantized version of 
FLUX.2-dev~\cite{bflFLUX2Frontier}. The quantized version is selected due to the standard model's extreme computation requirements.

Our models are conditioned on random 
test crops, whereas the FLUX.2 baseline uses full CLAHE-normalized test images paired 
with a domain-specific text prompt (further details describing the usage of FLUX.2-dev are present in the supplementary material). Generative 
quality is assessed via Fr\'echet Inception Distance (FID) and Kernel Inception 
Distance (KID) between the test set conditioning images and the generated images, using an Inception-v3~\cite{szegedy2016rethinking} backbone. The conditional adherence is measured by the cosine similarity of average-pooled DINOv2 tokens between 
reference and generated images. All metrics are computed at BenthicFlow's CFM standard training resolution ($224\times224$).

\begin{figure}
    \centering
    \includegraphics[width=0.91\linewidth]{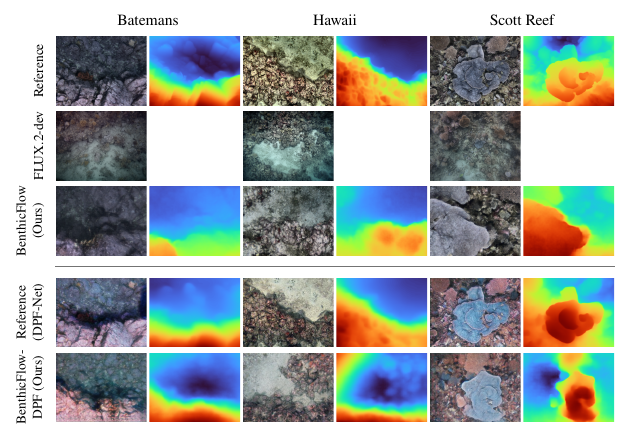}
    \caption{Qualitative comparison of the generative quality of the BenthicFlow models compared to FLUX.2-dev. Each column pair contains the output of a different reference image, with the left column containing the image and the right its predicted depth. The depth column is left blank for FLUX.2-dev, as it does not generate that modality.}
    \label{fig:qualitative-comparison}
    \vspace{-10pt}
\end{figure}

As shown in \cref{tab:generation_quality_regional},  BenthicFlow closely matches the test data  
distribution, achieving an FID $\le 20$ across all reefs and preprocessing 
variants (a result also present in the detailed per-campaign tests available in the supplementary material). This consistent high fidelity across diverse reefs demonstrates that 
our framework effectively models multi-modal distributions without sample 
degradation. Compared to FLUX.2-dev, BenthicFlow yields superior fidelity and 
higher conditional similarity. Qualitatively,  \cref{fig:qualitative-comparison} demonstrates how FLUX.2 introduces structural and optical artifacts that alter the identity of the target biome, whereas BenthicFlow preserves the visual identity of the reference reef while maintaining the natural environmental variations. Moreover, while the depth maps of the generative models appear slightly diffused compared to the references due to their lower resolution, they remain consistent with the generated images. 

To evaluate geographically discriminative retention, $k$-NN classifiers trained on average-pooled DINOv2 features predict the source reef (see \cref{tab:classification_consistency_complete}). BenthicFlow retains over 95\% of the balanced accuracy and macro F1 of real CLAHE data, while BenthicFlow-DPF retains 95\% and 94\% of real DPF-Net data; both substantially outperform FLUX.2-dev. These results indicate that our models preserve site-specific visual cues across distinct marine environments.

\begin{table}
\centering
\caption{Discriminative capacity evaluation using a $k$-NN classifier ($k=20$, $2,000$ training samples per reef). Classifiers are trained on either real data or synthetic data generated by the respective models, and evaluated on the test set. Reef columns report class-specific F1-scores (Scott refers to the Scott Reef location and Batemans abbreviates Batemans Bay), while global performance is captured via Balanced Accuracy (Bal. Acc.)~\cite{brodersen2010balanced} and Macro F1-score. Best results of the generative models are in \textbf{bold}.}
\label{tab:classification_consistency_complete}
\resizebox{0.8\linewidth}{!}{
\begin{tabular}{l@{\hspace{2em}}c@{\hspace{1.5em}}c@{\hspace{1.5em}}c@{\hspace{2em}}c@{\hspace{1.5em}}c}
\toprule
 & \multicolumn{3}{c}{\textbf{F1}} & & \\
\cmidrule(l{0em}r{2.05em}){2-4}
\textbf{Model} & \textbf{Batemans} & \textbf{Hawaii} & \textbf{Scott} & \textbf{Bal. Acc. $\uparrow$} & \textbf{Macro F1 $\uparrow$} \\
\midrule
\multicolumn{6}{@{}l}{\textit{Real Data}} \\
CLAHE & 0.928 & 0.834 & 0.850 & 0.873 & 0.871 \\
DPF-Net  & 0.910 & 0.783 & 0.869 & 0.865 & 0.854 \\
\midrule
\multicolumn{6}{@{}l}{\textit{Generated Data}} \\
FLUX.2-dev   & 0.260 & 0.667 & 0.697 & 0.568 & 0.541 \\
BenthicFlow         & \textbf{0.855} & \textbf{0.797} & 0.843 & \textbf{0.835} & \textbf{0.832} \\
BenthicFlow-DPF       & 0.841 & 0.710 & \textbf{0.853} & 0.823 & 0.802 \\
\bottomrule
\end{tabular}}
\vspace{-15pt}
\end{table}

\subsection{Extensible Generation and 3D Rendering}

\subsubsection{Window merging}
\label{window-merging}

The efficacy of our MultiDiffusion-inspired approach for generating extensible terrain 
depends heavily on the smooth integration of overlapping windows.  When visible seams are present in the image, they tend to appear in the center of the region where two windows overlap, since both windows have maximum contextual bias and equally strong weights (for a visual demonstration of this effect, see the supplementary material). To quantify the presence of these boundary artifacts, we 
introduce a targeted diagnostic tool denoted the Seam Gradient Ratio ($\text{SGR}$). Let $I$ denote the generated mosaic and let $S$ represent the 
sampling stride size used to generate it. We define $\Omega_{\text{seam}}$ as the localized spatial 
band centered along the window overlap boundary with a total width equal to half 
a stride ($S/2$), while $\Omega_{\text{rest}} = I \setminus \Omega_{\text{seam}}$ 
constitutes the remainder of the image canvas. The score evaluates the ratio of 
the average gradient magnitudes between these two domains:
\begin{equation}
\text{SGR} = \frac{\frac{1}{|\Omega_{\text{seam}}|} \sum_{(x,y) \in \Omega_{\text{seam}}} \|\nabla I(x, y)\|}{\frac{1}{|\Omega_{\text{rest}}|} \sum_{(x,y) \in \Omega_{\text{rest}}} \|\nabla I(x, y)\|}
\end{equation}
, where $\|\nabla I(x, y)\|$ represents the standard spatial gradient magnitude at 
pixel coordinates $(x,y)$. $\text{SGR}$ near $1$ indicates that the boundary 
region is indistinguishable from the rest of the generated texture, 
whereas a significantly different score indicates a sharp structural discontinuity or a visible 
seam. To ensure statistical soundness and account for natural texture-driven 
variations, this score is averaged across a large corpus of generated terrains.

We leverage this metric to systematically analyze how boundary seamlessness behaves 
as a function of window overlap. The evaluation consists of generating terrains with the standard BenthicFlow model conditioned on the test set across 7 overlapping configurations, ranging from 0\% (stride 16) to 87.5\% (stride 2). 500 images are generated for each ratio at a scale 16 times larger than the output resolution of the CFM. As illustrated in \cref{fig:stitching_gradient_ratio}, the gradient ratio is optimized between 50\% 
and 75\% overlap, indicating optimal blending at 62.5\%, which we adopt for further experiments. Although the ratio is the least optimal when there is no overlap, performance also decreases at high overlap percentages, such as 87.5\%.

\subsubsection{Latent interpolation}

\begin{figure}[t]
    \centering
    \begin{subfigure}[b]{0.46\textwidth}
        \centering
        \includegraphics[width=\textwidth]{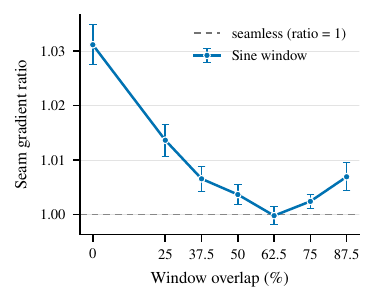}
        \caption{Seam gradient ratio metric across increasing overlaps. 95\% confidence intervals are represented with vertical lines.}
        \label{fig:stitching_gradient_ratio}
    \end{subfigure}
    \hfill
    \begin{subfigure}[b]{0.51\textwidth}
        \centering
        \includegraphics[width=\textwidth]{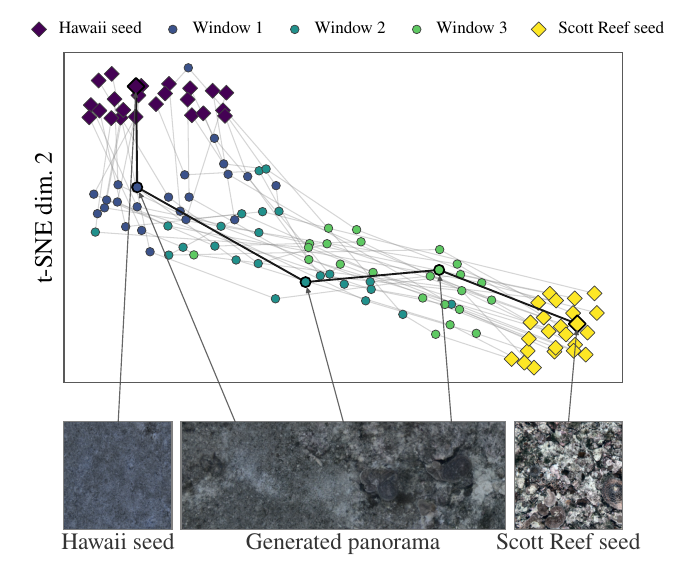}
        \caption{Interpolation of two reference images.}
        \label{fig:tsne-panorama}
    \end{subfigure}
    \vspace{0.2cm} 
    \caption{Analysis of the extensible generation strategy. (a) Visualization of the seam gradient ratio metric across different window overlaps. (b) t-SNE visualization of the interpolation windows between a set of  images, demonstrating progressive semantic evolution between references.}
    \label{fig:combined_generation_analysis}
    \vspace{-15pt}
\end{figure}

One of our assumptions about BenthicFlow's ability to generate diverse scenes is that a simple linear interpolation approach is sufficient to generate a smooth structural and semantic transition between image seeds. In \cref{fig:tsne-panorama}, we analyze this assumption by interpolating images of two different locations. To visualize the transition in semantic content of the panoramas, the DINOv2 patch tokens of the windows and the seeds in the two extremes are extracted and averaged, forming an embedding per window. The first 30 principal components of these averaged embeddings are extracted and their dimensionality is further reduced to two components using the t-SNE~\cite{van2008visualizing} algorithm. The resulting figure suggests a smooth transition from one window to the other in the embedding space, where closeness in this space significantly correlates with overlapping positions in the panorama.  

    
    
    
    

\subsubsection{Depth Consistency}

As a 2.5D image generative model, BenthicFlow must be capable of generating highly consistent texture and depth map pairs. A natural evaluation approach would involve computing the monocular depth estimation of the generated images and comparing it directly to their corresponding generated depth maps. However, because the generative models in this work are trained to produce localized image crops and DAv2 is highly sensitive to the loss of contextual cues in the periphery of the images, a direct comparison is not applicable.

To circumvent this, the consistency between the generated depth and image channels is approximated via the d-RAE's capacity to encode and decode image-depth pairs on unseen test data. The underlying premise is that if the d-RAE decoder generalizes to unseen data, it will similarly generalize to the unseen latents generated by our flow matching procedure. Furthermore, because full-sized test images can be processed by DAv2 prior to cropping, this evaluation bypasses the context-deficit limitation described above, enabling a more accurate assessment of the generator's consistency. As shown in \cref{tab:depth_consistency}, the metrics demonstrate a near-perfect alignment between the input and reconstructed depth, validating the structural and cross-modal consistency of BenthicFlow.

\begin{table}
\centering
\caption{Depth generation consistency evaluated against Depth Anything V2 pseudo-ground truth on the test dataset using root mean squared error (RMSE) and Pearson correlation. Best results in \textbf{bold}.}
\label{tab:depth_consistency}
\resizebox{0.5\linewidth}{!}{
\begin{tabular}{@{}lcc@{}}
\toprule
\textbf{Model} & \textbf{RMSE $\downarrow$} & \textbf{Pearson $r$ $\uparrow$} \\
\midrule
BenthicFlow d-RAE           & \textbf{0.006} & \textbf{0.997} \\
BenthicFlow-DPF d-RAE         & 0.009 & 0.992 \\
\bottomrule
\end{tabular}}
\vspace{-5pt}
\end{table}

\subsubsection{3D Rendering}

\begin{figure}
    \centering
    \includegraphics[width=0.95\linewidth]{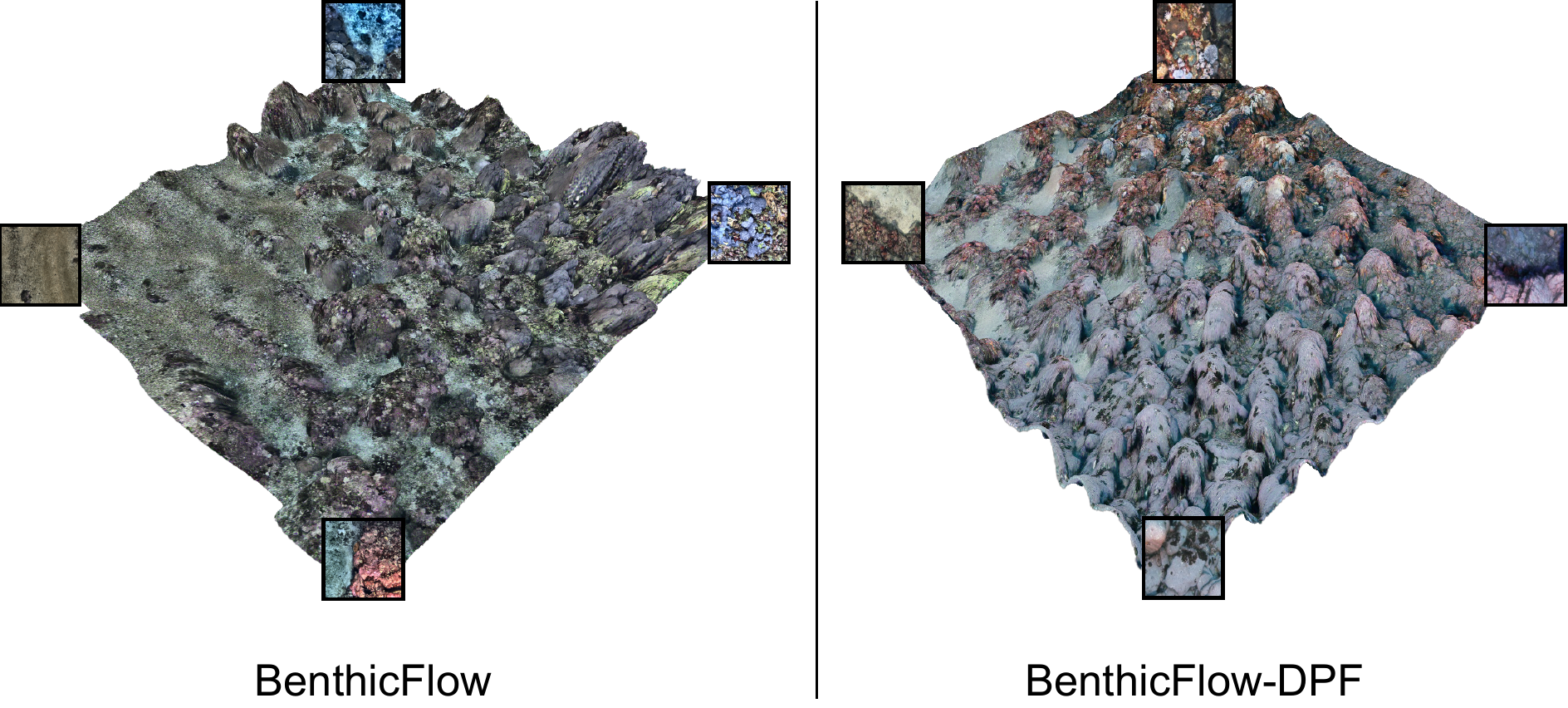}
    \caption{Two underwater scenes synthesized with BenthicFlow (left) and BenthicFlow-DPF (right) at a scale 32 times larger than the CFM's training resolution. The reference seeds used to render them are visible within frames on the corners of the scenes. }
    \label{fig:3d-rendering}
    \vspace{-15pt}
\end{figure}

\Cref{fig:3d-rendering} demonstrates the final output of the full generative pipeline. Both BenthicFlow variants successfully interpolate reference images from different geographic reefs to construct coherent, diverse landscapes. For example, the scene on the left shows how sand from the leftmost reference transitions into the rocky structures from the other references, and it is populated with realistically distributed plating corals and algae. The scene on the right demonstrates how the integration of DPF-Net~\cite{mei_dpf-net_2025} mitigates the underwater optical attenuation, yielding a larger color range and greater texture quality. Additional generated scenes are provided in the supplementary material.

To quantitatively evaluate the 3D lifting stage, we ablate both the base primitive (surface-aligned surfel vs.\ isotropic 3D Gaussian) and the appearance refinement strategy. Using $300$ RGBD samples generated from held-out \emph{test} deployments, we unproject the mosaic, build the primitives, optionally apply refinement, and render the front view against the generated reference. Since all configurations share identical inputs, comparisons are strictly paired. Note that this protocol measures single-view fidelity and does not quantify the off-axis multi-view coherence that motivates the use of surfels (Sec.~\ref{sec:lifting}).

As shown in \cref{tab:splat_ablation}, anchored refinement provides the largest improvement in source-view fidelity. Optimizing DC color, opacity, and bounded in-plane scale while fixing positions and orientations reduces LPIPS from $0.138$ to $0.046$ and increases SSIM from $0.884$ to $0.925$, improving LPIPS on 94\% of samples without allowing primitives to expand into volumetric blobs. Surface-aligned initialization also outperforms isotropic initialization before refinement across all metrics, showing that the geometric prior is beneficial independently of appearance fitting. Perceptual color polishing provides no further reprojection benefit and slightly degrades performance. Although isotropic Gaussians achieve a $0.39$~dB higher PSNR after convergence through greater volumetric flexibility, surfels preserve the surface alignment needed for multi-view coherence at only a minor single-view cost. We therefore adopt surface-aligned surfels with anchored refinement and no perceptual polish.

\begin{table}[t]
\centering
\caption{Lifting-stage ablation over $300$ generated samples (front-view
reprojection vs.\ reference; paired, single view). We report Peak Signal to Noise Ratio (PSNR), Structural Similarity Index Measure (SSIM) and LPIPS. The three components of the lifting-stage are ablated: the use of surfels (Surf.), anchored refinement (Ref.) and perceptual color polishing (Perc.). 
Best per column in \textbf{bold}.}
\label{tab:splat_ablation}
\setlength{\tabcolsep}{5pt}
\resizebox{0.5\linewidth}{!}{
\begin{tabular}{@{}ccc@{\hspace{10pt}}ccc@{}}
\toprule
\textbf{Surf.} & \textbf{Ref.} & \textbf{Perc.} & \textbf{PSNR $\uparrow$} & \textbf{SSIM $\uparrow$} & \textbf{LPIPS $\downarrow$} \\
\midrule
\xmark & \xmark & \xmark & 29.20 & 0.863 & 0.149 \\
\cmark & \xmark & \xmark & 30.03 & 0.884 & 0.138 \\
\cmark & \cmark & \xmark & 30.96 & 0.925 & \textbf{0.046} \\
\cmark & \cmark & \cmark & 30.94 & 0.925 & 0.048 \\
\xmark & \cmark & \cmark & \textbf{31.33} & \textbf{0.927} & 0.045 \\
\bottomrule
\end{tabular}}
\vspace{-10pt}
\end{table}


\section{Conclusions and Future Work}

This work introduces a unified framework for modeling diverse marine biomes and generating extensible 2.5D underwater environments. We show that Mul\-tiDiffusion-inspired sampling can be integrated directly into a flow-matching trajectory, providing a simpler alternative to terrain-generation pipelines that require a separately trained inpainting model. We further evaluate the framework across image generation, depth consistency, extensible synthesis, and 3D reprojection fidelity. Surface-aligned Gaussian surfels mitigate the limitations of lifting single-view RGBD mosaics by improving surface coverage while preserving the estimated geometry. Together, these components provide a scalable basis for generating synthetic underwater environments for downstream perception and robotics applications.

Several limitations motivate future work. First, the current conditioning uses average-pooled DINOv2 patch tokens; alternative representations, including CLS tokens and intermediate-layer aggregation, may provide more informative appearance control. Second, the pipeline remains fundamentally 2.5D and cannot recover geometry absent from the generated view. Future work could investigate direct 3D generation using pseudo-3D supervision from models such as VGGT-$\Omega$~\cite{wang2026vggt}. Third, high-resolution generation is constrained by the memory cost of decoding the full latent canvas, motivating a consistent windowed-decoding strategy. Finally, replacing cropped training with full-image training at higher resolutions may improve global structure and retain finer details after 3D lifting.

\section*{Acknowledgements}
This work was carried out within the ITEA Advisor and Xecs Marisens projects.

\bibliographystyle{splncs04}
\bibliography{main}

\title{Supplementary Material for BenthicFlow: Generating Extensible Underwater Environments via Flow Matching}
\author{Joaquín Figueira\textsuperscript{*} \and 
Camile Lendering\textsuperscript{*} \and
Manfred Gonzalez-Hernandez \and
Giacomo D'Amicantonio \and
Erkut Akdag \and
Egor Bondarev}

\authorrunning{J.~Figueira, C.~Lendering et al.}

\institute{
AIMS Group, Department of Electrical Engineering, Eindhoven University of Technology,
Eindhoven, The Netherlands\\
\email{\{j.figueira,c.r.lendering\}@tue.nl}\\[2pt]
\textsuperscript{*}Equal contribution.
}
\maketitle

\setcounter{section}{0}
\setcounter{figure}{0}
\setcounter{table}{0}
\setcounter{equation}{0}

\renewcommand{\thesection}{S\arabic{section}}
\renewcommand{\thefigure}{S\arabic{figure}}
\renewcommand{\thetable}{S\arabic{table}}
\renewcommand{\theequation}{S\arabic{equation}}

\section{Additional Single Image Generation Results}
\label{sec:extended-single-image}

\begin{table}[ht!]
\centering
\caption{Per-campaign \textbf{d-RAE reconstruction} results. Generative fidelity metrics:
FID and KID (lower is better), DINOv2 cosine similarity (Cos. Sim.) of mean-pooled tokens (it measures semantic alignment, higher is better); reconstruction
is evaluated only for the two flow-based pipelines. Depth consistency metrics: RMSE
(lower is better) and Pearson correlation $r$ (higher is better) compare reconstructed to reference
depth. For each campaign, we highlight the best results between the two models in \textbf{bold}.}
\label{tab:supp_recon}
\small
\setlength{\tabcolsep}{5pt}
\resizebox{\textwidth}{!}{
\begin{tabular}{l l r r r@{\,$\pm$\,}l r r}
\toprule
Model & Campaign & FID~$\downarrow$ & Cos. Sim.~$\uparrow$ & \multicolumn{2}{c}{KID~$\downarrow$} & Depth RMSE~$\downarrow$ & Depth $r$~$\uparrow$ \\
\midrule
\multirow{8}{*}{BenthicFlow d-RAE}
 & Batemans201011  & 14.8 & 0.874 & 0.0117 & 0.0011 & 0.0044 & 0.997 \\
 & Batemans201211  & 17.3 & 0.876 & 0.0130 & 0.0010 & \textbf{0.0074} & \textbf{0.997} \\
 & Batemans201411  & 23.1 & 0.870 & 0.0216 & 0.0010 & \textbf{0.0041} & \textbf{0.999} \\
 & Hawaii201801    & 15.0 & \textbf{0.883} & 0.0095 & 0.0009 & \textbf{0.0052} & \textbf{0.998} \\
 & ScottReef200907 & \textbf{8.8}  & 0.929 & \textbf{0.0070} & 0.0008 & \textbf{0.0061} & \textbf{0.998} \\
 & ScottReef201108 & \textbf{7.1}  & 0.920 & \textbf{0.0049} & 0.0006 & \textbf{0.0072} & 0.997 \\
 & ScottReef201503 & 13.7 & 0.929 & \textbf{0.0070} & 0.0006 & \textbf{0.0077} & 0.997 \\
 & \textit{Average} & 14.3 & 0.891 & 0.0107 & 0.0021 & \textbf{0.0060} & \textbf{0.998} \\
\midrule
\multirow{8}{*}{BenthicFlow-DPF d-RAE}
 & Batemans201011  & \textbf{12.8} & \textbf{0.884} & \textbf{0.0084} & 0.0005 & \textbf{0.0043} & 0.979 \\
 & Batemans201211  & \textbf{15.3} & \textbf{0.880} & \textbf{0.0109} & 0.0009 & 0.0086 & 0.990 \\
 & Batemans201411  & \textbf{16.1} & \textbf{0.899} & \textbf{0.0136} & 0.0011 & 0.0063 & 0.997 \\
 & Hawaii201801    & \textbf{14.4} & 0.873 & 0.0095 & 0.0012 & 0.0120 & 0.998 \\
 & ScottReef200907 & 11.6 & \textbf{0.937} & 0.0114 & 0.0010 & 0.0064 & 0.982 \\
 & ScottReef201108 & 8.6  & \textbf{0.935} & 0.0061 & 0.0007 & 0.0123 & \textbf{0.998} \\
 & ScottReef201503 & \textbf{11.1} & \textbf{0.943} & 0.0074 & 0.0013 & 0.0130 & \textbf{0.998} \\
 & \textit{Average} & \textbf{12.8} & \textbf{0.907} & \textbf{0.0096} & 0.0010 & 0.0090 & 0.992 \\
\bottomrule
\end{tabular}}
\end{table}

Extended results for single image generation evaluations are provided in tables \cref{tab:supp_recon} and \cref{tab:supp_generation}. The tables contain results for every campaign available in the dataset. As seen in the table, the results remain consistent not only at the geographic location level but also at the campaign level. Additionally, these extended metrics show the relatively inferior performance of the Batemans campaigns. We hypothesize that this is caused by a moderate imbalance in the training data: the Scott Reef and Hawaii locations have more images per campaign available in the Squidle+~\cite{friedman2025squidle} framework, which leads to the flow matching models receiving a stronger learning signal for those locations.    

\begin{table}
\centering
\caption{Per-campaign \textbf{conditional generation} results. FID and KID measure
distributional fidelity (lower is better); DINOv2 cosine similarity (Cos. Sim.) of mean-pooled tokens measures semantic
alignment (higher is better). For each campaign the best results between the two models are shown in \textbf{bold}.}
\label{tab:supp_generation}
\small
\setlength{\tabcolsep}{6pt}
\begin{tabular}{l l r r r@{\,$\pm$\,}l}
\toprule
Model & Campaign & FID~$\downarrow$ & Cos. Sim.~$\uparrow$ & \multicolumn{2}{c}{KID~$\downarrow$} \\
\midrule
\multirow{8}{*}{BenthicFlow}
 & Batemans201011  & 27.1 & \textbf{0.817} & 0.0234 & 0.0017 \\
 & Batemans201211  & \textbf{16.0} & 0.797 & 0.0075 & 0.0007 \\
 & Batemans201411  & \textbf{16.9} & 0.801 & 0.0110 & 0.0008 \\
 & Hawaii201801    & \textbf{13.9} & \textbf{0.833} & \textbf{0.0070} & 0.0005 \\
 & ScottReef200907 & \textbf{8.4}  & 0.888 & \textbf{0.0046} & 0.0003 \\
 & ScottReef201108 & \textbf{9.0}  & 0.870 & \textbf{0.0043} & 0.0004 \\
 & ScottReef201503 & 22.2 & 0.888 & 0.0104 & 0.0007 \\
 & \textit{Average} & 16.2 & \textbf{0.842} & 0.0097 & 0.0026 \\
\midrule
\multirow{8}{*}{BenthicFlow-DPF}
 & Batemans201011  & \textbf{24.9} & 0.774 & \textbf{0.0174} & 0.0008 \\
 & Batemans201211  & 16.4 & \textbf{0.800} & 0.0075 & 0.0008 \\
 & Batemans201411  & 17.6 & \textbf{0.812} & \textbf{0.0108} & 0.0008 \\
 & Hawaii201801    & 15.5 & 0.804 & 0.0079 & 0.0010 \\
 & ScottReef200907 & 10.8 & \textbf{0.891} & 0.0080 & 0.0005 \\
 & ScottReef201108 & 9.2  & \textbf{0.883} & 0.0044 & 0.0005 \\
 & ScottReef201503 & \textbf{15.7} & \textbf{0.908} & \textbf{0.0079} & 0.0006 \\
 & \textit{Average} & \textbf{15.7} & 0.839 & \textbf{0.0091} & 0.0017 \\
\midrule
\multirow{8}{*}{FLUX.2-dev}
 & Batemans201011  & 142.4 & 0.489 & 0.1298 & 0.0027 \\
 & Batemans201211  & 100.9 & 0.612 & 0.0849 & 0.0011 \\
 & Batemans201411  & 87.5  & 0.617 & 0.0750 & 0.0012 \\
 & Hawaii201801    & 137.1 & 0.593 & 0.1247 & 0.0024 \\
 & ScottReef200907 & 75.2  & 0.765 & 0.0639 & 0.0016 \\
 & ScottReef201108 & 73.1  & 0.763 & 0.0579 & 0.0013 \\
 & ScottReef201503 & 81.9  & 0.707 & 0.0585 & 0.0012 \\
 & \textit{Average} & 99.7 & 0.649 & 0.0850 & 0.0017 \\
\bottomrule
\end{tabular}
\end{table}

\section{Extended Analysis of the Extensible Generation}
\label{sec:window-merging-qualitative}

\begin{figure}[ht!]
    \centering
    \begin{subfigure}[t]{0.48\textwidth}
        \centering
        \includegraphics[width=0.85\linewidth]{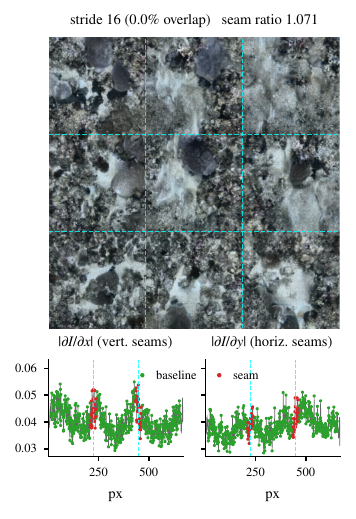}
        \caption{Above, an example of a terrain generated with non-overlapping windows. Below, the average gradient of each vertical and horizontal position in the image.}
        \label{fig:stitching-window-16}
    \end{subfigure}
    \hfill
    \begin{subfigure}[t]{0.48\textwidth}
        \centering
        \includegraphics[width=0.85\linewidth]{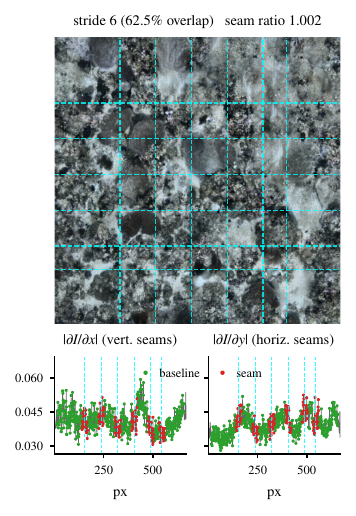}
        \caption{Above, a terrain generated with 62.5\% overlap. Below, the average gradient of each vertical and horizontal position in the image.}
        \label{fig:stitching-window-06}
    \end{subfigure}
    \caption{Two generated terrains and the variation of their gradient magnitudes in both axes. Cyan dashed lines are used to indicate the region of the image where we would expect noticeable seams to appear (\ie the center of the region where two windows overlap and where the pull of multiple trajectories equalizes).}
    \label{fig:stitching-windows}
\end{figure}

To qualitatively analyze the effects of different overlap percentages, we present in \cref{fig:stitching-windows} two examples of terrains generated using the standard BenthicFlow model. They consist of mosaics 16 times larger than BenthicFlow's CFM standard output (4 times larger vertically and 4 times larger horizontally). In \cref{fig:stitching-window-16}, where no overlapping windows were applied, clear discontinuities can be observed between the windows, especially in terms of lighting and color consistency. This is also observable in the plot of the gradients below the figure, where clear relative peaks can be observed near the regions of the seams labeled with cyan dashed lines. In contrast, the image in \cref{fig:stitching-window-06} shows that there are no visible discontinuities or abrupt color changes between windows when using the optimal overlap range. Moreover, there are few visible peaks in the plot of the gradients where seams would be expected (dashed blue lines), indicating a more uniform gradient profile surrounding seam heavy areas. Note also the difference between SGR scores between the two images: the non-overlapping terrain has an SGR of 1.07, while the optimal overlapping percentage has a score of 1.002, significantly closer to the perfect ratio of 1.

\begin{figure}[ht!]
    \centering
    \includegraphics[width=0.5\linewidth]{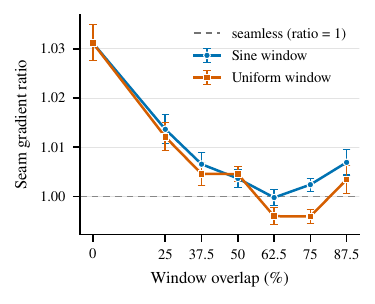}
    \caption{Comparison of the seam gradient ratio between employing a uniform weight window and the standard sine window of BenthicFlow.}
    \label{fig:stitching-seam-ratio-both}
\end{figure}

A final additional result motivating our use of a sine window merging strategy can be seen in \cref{fig:stitching-seam-ratio-both}. The figure exhibits the SGR score as a function of the degree of overlap using two window averaging strategies: a uniformly weighted window and our adopted sine weight window. We can see that, although the weight configuration has less importance in the extremes of the overlap range, it is only possible to obtain an optimal ratio using the sine window function.  

\section{Benthic Conditioning of FLUX.2-dev}
\label{sec:flux-details}

We condition FLUX.2-dev using a reference $512\times512$ image and with the following prompt, which encompasses the main properties of the benthic images considered in this work, while simultaneously encouraging output diversity: 

\textit{``Top-down nadir benthic survey photo of the underwater sea floor. Flat orthographic perspective with artificial strobe lighting. Substrate showing hard corals, rubble, and sand. Very slight water attenuation and marine snow backscatter. Scientific documentary style. Change the distribution of the benthic substrate, but keep the same perspective, lighting, and style''}.

A final detail related to our usage of FLUX.2-dev is the number of diffusion steps employed during inference. We decide to use 28 steps as suggested in \cite{huggingfaceBlackforestlabsFLUX2devHugging}, which provides an appropriate trade-off between image quality and inference speed. Further motivation for this decision, which also drives our use for the quantized version of FLUX.2-dev, is to equalize computational resources: 28 steps of FLUX.2-dev 4-bit quantized on an H100 NVIDIA GPU generate one image every $\sim$ 6 seconds, while BenthicFlow is able to generate more than 120 images on the same time frame while employing the same hardware. 

\section{Additional Generated Scenes}
\label{sec:additional-scenes}

\begin{figure}
    \centering
    \includegraphics[width=1.0\linewidth]{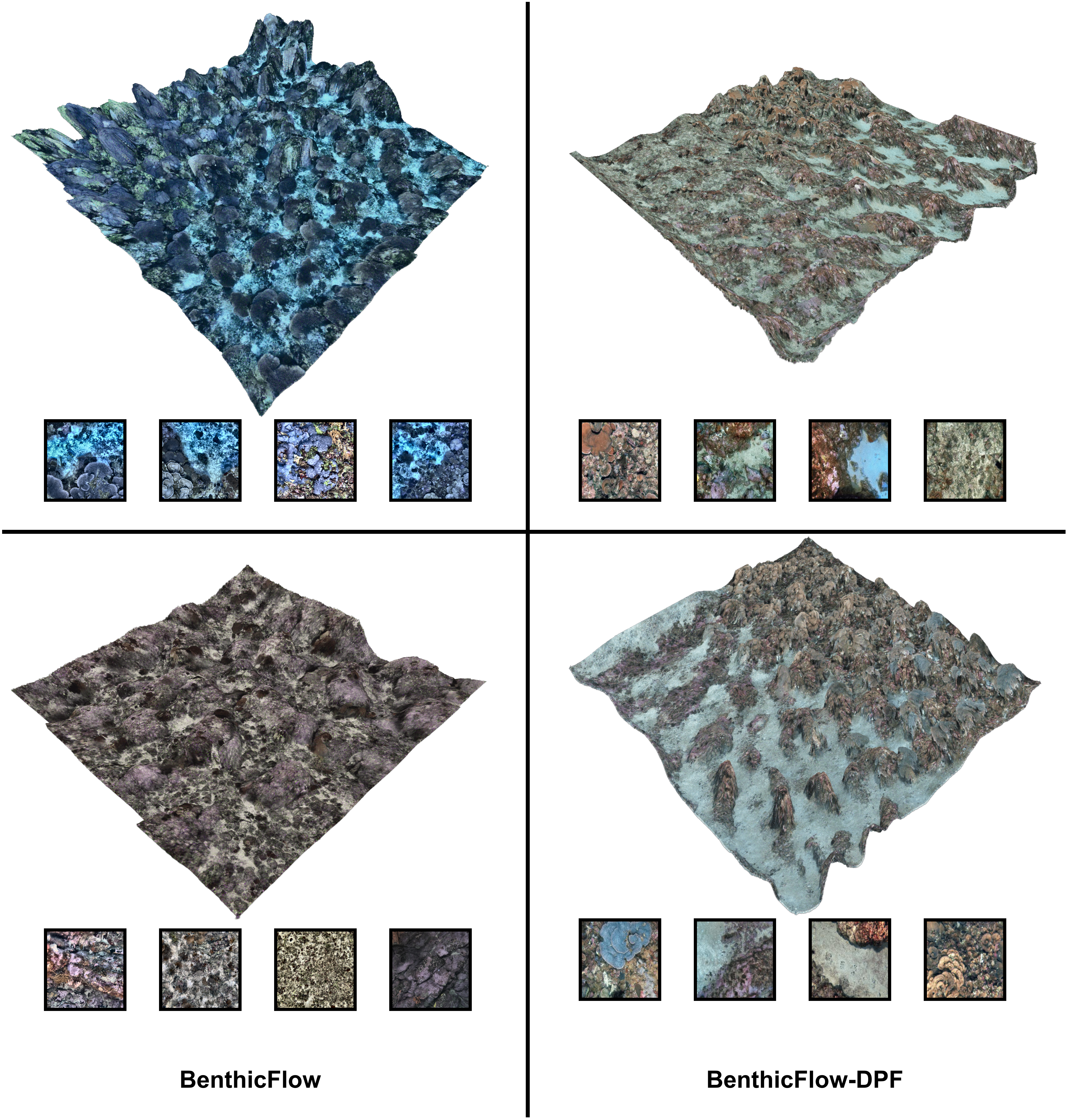}
    \caption{Additional generated scenes for BenthicFlow (left) and BenthicFlow-DPF (right). All reference seeds, visible in frames below the generated scenes, are sampled from different deployments and campaigns.}
    \label{fig:additional-scenes}
\end{figure}

To further demonstrate BenthicFlow's ability to generate diverse scenes, we provide additional 3D renderings generated at a scale 36 times larger than the model's window resolution (6 times larger horizontally, and 6 times larger vertically). These scenes can be observed in \cref{fig:additional-scenes}, demonstrating smooth transitions between the seeds and natural benthic biome variations. 

\section{Dataset Details}
\label{sec:dataset-details}

In order to ensure reproducibility of our results, we provide the specific subset of the Squidle+ framework~\cite{friedman2025squidle} we selected for our experiments. The selection, visible in \cref{tab:data_splits}, is provided at the deployment level. The split and campaign of each deployment are also indicated.

\begingroup
  \scriptsize
  \setlength{\tabcolsep}{4pt}
  \setlength{\LTcapwidth}{\linewidth}
  \begin{longtable}{@{}lllr@{}}
    \caption{Deployment-level data split. Each AUV campaign is split by
    \emph{whole deployments}: the validation and test sets each hold out
    one complete deployment per campaign, never seen during training, so
    no spatially overlapping imagery leaks across splits. Totals:
    510,307 train / 76,480 val / 87,574 test images
    over 52/7/7 deployments.}
    \label{tab:data_splits}\\
    \toprule
    Campaign & Split & Deployment & \#Images \\
    \midrule
    \endfirsthead
    \multicolumn{4}{@{}l}{\tablename~\thetable{} -- continued from previous page}\\[2pt]
    \toprule
    Campaign & Split & Deployment & \#Images \\
    \midrule
    \endhead
    \multicolumn{4}{r@{}}{continued on next page}\\
    \endfoot
    \bottomrule
    \endlastfoot
    Batemans201011 & Train & \texttt{r20101117\_001912\_batemans\_03\_site1sz} & 8,872 \\*
     &  & \texttt{r20101117\_224623\_batemans\_06\_site6guz} & 3,165 \\*
     &  & \texttt{r20101118\_010326\_site6guz\_08\_broadgrid} & 10,011 \\*
     &  & \texttt{r20101118\_053354\_site4sz\_09\_transect} & 4,188 \\*
     &  & \texttt{r20101119\_224028\_site6guz\_11\_densegrid3} & 2,411 \\*
     &  & \texttt{r20101120\_002302\_site5sz\_12\_densegrids} & 8,486 \\*
     &  & \texttt{r20101120\_192136\_site3guz\_13\_densegrids} & 8,855 \\*
     &  & \texttt{r20101120\_222627\_site2guz\_14\_densegrids} & 8,145 \\*
     &  & \texttt{r20101121\_062424\_site1sz\_16\_densegrids3} & 2,776 \\*
     &  & \texttt{r20101122\_185420\_site4sz\_17\_densegrids} & 8,311 \\*
     &  & \texttt{r20101122\_214134\_site4sz\_18\_broadgrid} & 11,211 \\*
     & Val & \texttt{r20101121\_030956\_site1sz\_15\_densegrids} & 5,893 \\*
     & Test & \texttt{r20101119\_204443\_site6guz\_10\_densegrids} & 6,554 \\
    \midrule
    Batemans201211 & Train & \texttt{r20121128\_025727\_Tollgates\_site4sz\_03\_dense} & 7,882 \\*
     &  & \texttt{r20121128\_051532\_Tollgates\_site4sz\_04\_broad} & 7,086 \\*
     &  & \texttt{r20121128\_195806\_Burrewarra\_BU\_DG3\_05\_dense} & 8,195 \\*
     &  & \texttt{r20121128\_225241\_Lilli\_Pilli\_LP\_DG2\_06\_dense} & 8,230 \\*
     &  & \texttt{r20121129\_011610\_Lilli\_Pilli\_LP\_DG2\_07\_broad} & 2,036 \\*
     & Val & \texttt{r20121127\_231520\_Durras\_site3guz\_02\_broad} & 9,764 \\*
     & Test & \texttt{r20121127\_204111\_Durras\_site3guz\_01\_dense} & 8,593 \\
    \midrule
    Batemans201411 & Train & \texttt{r20141111\_220936\_01\_Tollgates\_site4sz\_broad} & 6,772 \\*
     &  & \texttt{r20141112\_004500\_02\_Tollgates\_site4sz\_dense} & 7,704 \\*
     &  & \texttt{r20141116\_202733\_03\_Durras\_site3guz\_broad} & 10,463 \\*
     &  & \texttt{r20141117\_030254\_05\_Burrewarra\_broad} & 10,291 \\*
     &  & \texttt{r20141117\_070259\_07\_Burrewarra\_BU\_DG3\_dense\_continue} & 3,823 \\*
     &  & \texttt{r20141117\_213857\_09\_Burrewarra\_BU\_DG3\_dense\_continue\_grid3} & 4,244 \\*
     &  & \texttt{r20141118\_012715\_11\_Lilli\_Pilli\_LP\_DG2\_dense} & 12,371 \\*
     & Val & \texttt{r20141116\_225628\_04\_Durras\_site3guz\_dense} & 12,525 \\*
     & Test & \texttt{r20141117\_053229\_06\_Burrewarra\_BU\_DG3\_dense} & 6,404 \\
    \midrule
    Hawaii201801 & Train & \texttt{r20180202\_195659\_SS11\_waikoloa\_broad\_cros\_200\_150} & 35,466 \\*
     &  & \texttt{r20180204\_190219\_SS13\_waikoloa\_legs\_deep} & 27,372 \\*
     &  & \texttt{r20180206\_012603\_SS15\_waikoloa\_legs\_mid\_central} & 18,021 \\*
     &  & \texttt{r20180206\_184219\_SS16\_waikoloa\_legs\_mid\_south} & 14,808 \\*
     &  & \texttt{r20180207\_010805\_SS17\_waikoloa\_auto\_goto} & 6,653 \\*
     & Val & \texttt{r20180205\_180015\_SS14\_waikoloa\_legs\_deep\_south} & 22,180 \\*
     & Test & \texttt{r20180203\_182731\_SS12\_waikoloa\_broad\_cross\_200\_150\_c\_shallow} & 34,609 \\
    \midrule
    ScottReef200907 & Train & \texttt{r20090726\_074343\_scott\_05\_long\_transect\_auv2} & 12,648 \\*
     &  & \texttt{r20090727\_002236\_scott\_06\_long\_transect\_auv5} & 15,533 \\*
     &  & \texttt{r20090727\_085810\_scott\_07\_grids\_auv2} & 12,775 \\*
     &  & \texttt{r20090729\_003637\_scott\_10\_long\_transect\_auv3} & 15,276 \\*
     &  & \texttt{r20090729\_072005\_scott\_11\_dense\_survey\_auv5\_deep} & 6,993 \\*
     &  & \texttt{r20090730\_001654\_scott\_13\_long\_transect\_auv6} & 11,188 \\*
     & Val & \texttt{r20090728\_074740\_scott\_09\_long\_transect\_auv1} & 14,862 \\*
     & Test & \texttt{r20090728\_004240\_scott\_08\_long\_transect\_auv4} & 10,725 \\
    \midrule
    ScottReef201108 & Train & \texttt{r20110808\_052328\_01\_scott\_grids\_deep\_auv1} & 462 \\*
     &  & \texttt{r20110810\_042127\_04\_scott\_long\_leg\_auv8} & 180 \\*
     &  & \texttt{r20110811\_222428\_08\_scott\_grids\_deep\_auv2} & 13,973 \\*
     &  & \texttt{r20110812\_042524\_09\_scott\_long\_leg\_auv2} & 12,781 \\*
     &  & \texttt{r20110812\_094330\_10\_scott\_grids\_deep\_auv3} & 15,644 \\*
     &  & \texttt{r20110812\_223024\_11\_scott\_grids\_shallow\_auv3} & 13,321 \\*
     &  & \texttt{r20110813\_223332\_14\_scott\_grids\_shallow\_auv5} & 13,179 \\*
     &  & \texttt{r20110814\_022231\_15\_scott\_long\_leg\_auv5} & 13,219 \\*
     & Val & \texttt{r20110808\_091108\_01a\_scott\_dense\_deep\_auv1} & 1,717 \\*
     & Test & \texttt{r20110813\_022516\_12\_scott\_long\_leg\_auv3} & 15,523 \\
    \midrule
    ScottReef201503 & Train & \texttt{r20150326\_071126\_01\_scott\_grids\_shallow\_auv4} & 7,901 \\*
     &  & \texttt{r20150327\_015552\_02\_scott\_grids\_deep\_auv4} & 14,109 \\*
     &  & \texttt{r20150328\_000850\_03\_scott\_grids\_deep\_auv4\_shifted} & 13,644 \\*
     &  & \texttt{r20150328\_042551\_04\_scott\_dense\_grid\_deep\_auv4} & 2,519 \\*
     &  & \texttt{r20150329\_005815\_05\_scott\_grids\_shallow\_auv5} & 5,300 \\*
     &  & \texttt{r20150329\_061201\_06\_scott\_long\_leg\_auv5} & 386 \\*
     &  & \texttt{r20150329\_231742\_07\_scott\_grids\_deep\_auv3} & 14,537 \\*
     &  & \texttt{r20150330\_035644\_08\_scott\_grids\_deep\_auv3\_rep\_dense\_grid} & 3,527 \\*
     &  & \texttt{r20150330\_225013\_09\_scott\_grids\_deep\_auv2} & 13,042 \\*
     &  & \texttt{r20150331\_050931\_10\_scott\_long\_leg\_auv2} & 10,322 \\*
     & Val & \texttt{r20150329\_063007\_06\_scott\_long\_leg\_auv5} & 9,539 \\*
     & Test & \texttt{r20150331\_231619\_11\_scott\_repeat\_large\_200907\_25\_auv5} & 5,166 \\
  \end{longtable}
\endgroup

\pagebreak


\end{document}